\documentclass[11pt]{article}
\usepackage{acl2014}
\usepackage{times}
\usepackage{url}
\usepackage{latexsym}

\usepackage{graphics}

\usepackage{algorithm}
\usepackage{algorithmic}

\usepackage{verbatim}

\usepackage[abs]{overpic}

\newcommand{\wt}{{\bf w_t}}

\newcommand{\z}[1]{{\bf z_{#1}}}
\newcommand{\s}{{\bf s}}

\newcommand{\g}{{\bf g}}

\newcommand{\ltopic}{{\it w-topic}}
\newcommand{\ltopics}{{\it w-topics}}

\newcommand{\htopic}{{\it t-topic}}
\newcommand{\htopics}{{\it t-topics}}

\renewcommand{\baselinestretch}{0.98}

\title{Coarse-grained Cross-lingual Alignment of Comparable Texts \\ with Topic Models and Encyclopedic Knowledge}

\author{ Vivi Nastase \\
  HLT group \\
  FBK, Trento, Italy \\
  nastase@fbk.eu \\ 
\And
  Angela Fahrni \\
  HITS gGmbH \\
  Heidelberg, Germany \\
  angela.fahrni@h-its.org }

\date{}

\begin{document}
\maketitle
\begin{abstract}
  We present a method for coarse-grained cross-lingual alignment of
  comparable texts: segments consisting of contiguous paragraphs that
  discuss the same theme (e.g. history, economy) are aligned based on
  induced multilingual topics. The method combines three ideas: a two
  level LDA model that filters out words that do not convey themes, an
  HMM that models the ordering of themes in the collection of
  documents, and language-independent concept annotations to serve as
  a cross-language bridge and to strengthen the connection between
  paragraphs in the same segment through concept relations. The method
  is evaluated on English and French data previously used for
  monolingual alignment. The results show state-of-the-art performance
  in both monolingual and cross-lingual settings.
\end{abstract}

\section{Introduction}

\begin{figure*}[t]
  \centering
  \includegraphics[scale=0.6]{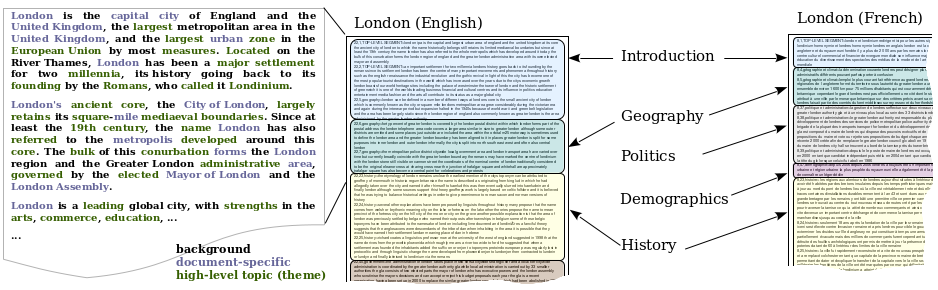}
  \vspace{-5mm}
  \caption{A two level -- paragraph and document -- topic modeling for cross-document and cross-lingual topic-based alignment. Within a paragraph, words pertaining to different topics are highlighted with different colours.} 
  \label{fig:overview}
\end{figure*}

Coarse-grained alignment of documents groups text segments in
different documents that convey the same theme -- e.g. {\it history,
  geography} in texts about cities. Cross-language alignment deals
with the added challenge of aligning segments from documents in
different languages. This could be useful as a prelude to fine-grained
alignment, or for building coarsely aligned multilingual corpora for
machine translation or text categorization. We are interested in
cross-lingual alignment for the synchronization of web (wiki) pages
with multiple language versions, where pages in different languages
are independently edited. A coarse alignment would (i) reveal quickly
text portions that are not shared, and must thus be translated and
added, and (ii) show potentially parallel portions, to be further
processed to produce a more fine-grained alignment. We present a
cross-lingual alignment of segments -- consisting of one or more
paragraphs -- based on knowledge-enhanced topic modeling, illustrated
in Figure \ref{fig:overview}, which implements the following ideas:

\paragraph{2-level LDA} A two-layer modeling framework (i) models paragraphs as a mixture of three topics -- background, document-specific and theme-specific (e.g. in a document about the city of Montreal, articles, prepositions, etc.\ would be background, words associated with Montreal but not specific to a theme -- Montreal, Canada, French, Quebec -- would be document-specific, and others such as hockey, sports, skating would be theme-specific); (ii) models a document as a mixture of K topics/themes.

\paragraph{HMM theme sequences} Following the assumption that themes are not presented randomly in a document, we use an HMM to model their sequence. State transitions are modeled through a Dirichlet distribution, tuned to bias the system towards self-transitions and thus avoid rapid switching between states \cite{beal02,teh06,fox10}.

\paragraph{Language-independent concept annotations} 
We inject knowledge in the model by linking single and multi-word terms
to concepts in a concept network obtained from Wikipedia
\cite{nastase13}. Terms are replaced with
language-independent identifiers (e.g. ``New York'' becomes
``c553795''). Enriching text with concept annotations allows to: (i)
integrate naturally the identified multi-word expressions into the
statistical process; (ii) deal with ambiguity (``New York'' the
city becomes ``c553795'', while ``New York'' the state becomes
``c27491610''); (iii) deal partly with coreferent mentions and
synonymy (``New York City'', ``The Big Apple'', ``Nueva York'' share
the same identifier); (iv) bridge languages and build
multilingual topic models, as concepts are shared across languages;
(v) use additional encyclopedic knowledge extracted from the
concept network, e.g.\ relations between concepts within the same or
consecutive paragraphs to strengthen the cohesion of the topic.

\paragraph{}The method is applied on Wikipedia articles about cities in English and French \cite{chen09}. This allows for comparison with related work (in the monolingual setting), and to extend the evaluation to a cross-lingual setting. The monolingual segment alignment is evaluated against a Hidden Topic Markov Model \cite{gruber07} -- and the GMM model \cite{chen09}. The method described here scores higher than both.
In the cross-lingual setting, the baselines are the alignments produced using translation tables, and using concept annotations, respectively. The method which combines topic modeling with concept annotations outperforms both baselines.

\section{Previous Work}\label{sec:previouswork}

\paragraph{Text alignment} 

For monolingual comparable corpora, text relatedness measures produce good alignment results \cite{barzilay03b,yahyaei11}. Methods for sentence alignment in a parallel bilingual corpus typically rely on linearity of the alignment, existence of 1:1 mapping between aligned sentences and the correlation of sentence length \cite{tiedemann11}. These assumptions do not hold for comparable (but not parallel) corpora. Bilingual comparable corpora have been mainly used to extract parallel fragments \cite{gupta13} or paraphrases and word translations \cite{fung97}. We focus on an alignment of fragments that have the same theme, but are not necessarily parallel.



\paragraph{Sequence modeling}

 Modeling a document as a sequence of topics brings HMMs naturally to mind. Work in this area started with \newcite{mulbregt98} and \newcite{blei01}. This approach suffers from two shortcomings: the number of states needs to be fixed (to be able to perform the Baum-Welch or Viterbi algorithms), and the HMM tends to switch fast between different states. In situations where a state should be persistent, as in topic segmentation, this is a problem. A breakthrough has come when state transitions in an HMM were modeled through a Dirichlet process \cite{beal02}. This allows one to control state transitions, and to let the model induce the number of states that best fit the data. \newcite{beal02} used three hyperparameters to control the HMM: for self-transitions, for transitions to previously used states, and for adding a new transition. \newcite{teh06} and \newcite{fox10} extended this model and showed how to manipulate parameters to avoid rapid switching between states.

\paragraph{Language models}
 In topic segmentation one can consider the
words to be the basic unit \cite{gruber07}, or sentences/paragraphs (e.g. \cite{blei01}, \cite{eisenstein08}). Sentences are themselves
heterogeneous. \newcite{daume06} used topic models for query-driven
summarization. The assumption is that each sentence consists of a
mixture of language models -- one corresponding to the general model
of the English language, one corresponding to the overall theme of the
document, and one that matches the topic of the given
query. \newcite{zhai04}, \newcite{titov08} and \newcite{paul09} model
topics (aspects) that run throughout a collection of documents, and
topics that are collection-specific. Considering a sentence/paragraph
as a small document, and one document as a collection of these small
documents, it could be construed that the previously mentioned work models a sentence/paragraph as
a mixture of topics, some of which are common throughout the document,
some of which are sentence-specific. \newcite{wang11} also assume a
sentence to be generated from a mixture of topics, in particular two: a ``functional'' (i.e. background) topic, and a content topic. Additionally, this approach models transitions between topics through an HMM, but like previous work, without concerns about the rapid state switching. This does not pose a problem because of the nature of the data: short ads and reviews with short sentences and not much topic repetition.

\paragraph{WSD and topic modeling} 
\newcite{boydgraber07} combine topic modeling with word sense disambiguation relative to WordNet, following the observation that words in different topics may exhibit different senses, and that the sense of a word depends on the context in which it appears. Each noun that appears in WordNet will enhance the probabilities of all paths from the root to each possible sense, and ``correct'' paths will have higher probability because of aggregated evidence from the words in the text. A downside of this is that, as topic models do, it deals with single word terms only, whereas texts contain numerous multi-word terms referring to real world entities. For this reason we apply the concept and entity identification process before the topic modeling step.

\paragraph{Multilingual topic modeling} \newcite{jagarlamudi10} use a bilingual dictionary to obtain multilingual topics from an unaligned multilingual corpus. They assume topics to be formed of concepts, which can have different lexicalizations depending on the language. These concepts consist of entries from a bilingual dictionary. \newcite{boydgraber09} simultaneously discover matching words and multilingual topics from a collection composed of documents in two languages, based on the assumption that similar words appear in similar contexts. \newcite{zhang10} use a bilingual dictionary as a source of constraints for bridging texts in two languages, using the assumption that related words in different languages have similar distributions. The topic models produced contain words in different languages that are allowed to have different probabilities, reflecting the difference between the data in the two languages. \newcite{ni09} mine multilingual topics from Wikipedia, using the articles on the same theme in different languages as a source of language models. \newcite{mimno09} build polylingual topic models from sets of documents on the same topics, in particular Wikipedia articles in several languages. They assume that the different language versions of an article have similar topic distributions, and topics consist of language-specific word distribution. \newcite{vulic11} build bilingual topic models for information retrieval, using comparable corpora -- in particular collections of Wikipedia articles on the same topics in the targeted languages. 


\section{A Topic Model for Alignment}\label{sec:model}
We build upon some of the ideas presented above to develop a knowledge-enhanced Hidden Topic Markov Model (HTMM): documents, made up of paragraphs, are modeled as sequences of topics, with transition between states controlled by a Dirichlet distribution. The process is detailed below.

\subsection{Overall generative process}

We describe here the overall generative process, covering the two level LDA and HMM modeling, which will be detailed separately below.

Assuming $K$ \htopics~(themes) represented in the documents $d_{1:M}$ in our collection, the generative process is as follows:
{\small
\begin{enumerate}
 \item[1] Draw a word distribution $\xi_1$ $\sim$ Dirichlet($\eta_1$) for a background language model;
 \item[2] Draw a word distribution $\phi_z$ $\sim$ Dirichlet($\beta$) for each \htopic~ $z \in \{1..K\}$;
 \item[3] draw a \htopic~transition probability distribution $\pi$ $\sim$ Dirichlet($\alpha+\kappa$) ($\pi_0$ is an initial state probability vector, $\pi_j$ is a transition probability distribution from state $j$);
 \item[4] For each document $d_m, m = 1..M $:
   \begin{enumerate}
     \item[4.1] draw a word distribution $\xi_{2m}$ $\sim$ Dirichlet($\eta_2$) for a document-specific language model;
     \item[4.2] draw a \htopic~mixture $\theta_m$ $\sim$ Dirichlet($\lambda$) for document $d_m$;
     \item[4.3] for each paragraph (sequence of words) $\wt_m$ in $d_m$:
       \begin{enumerate}
         \item[4.3.1] sample a \htopic~$z_t$ $\sim$ Discrete($\pi_{z_{t-1}}, \theta_m$);
         \item[4.3.2] draw a \ltopic~mixture $\psi_m$ $\sim$ Dirichlet($\gamma$);
         \item[4.3.3] for each word $w_{ti}$ in sequence $\wt_m$:
           \begin{enumerate}
             \item sample a topic $s_{ti}$ $\sim$ Discrete($\psi_m$);
             \item if $s_{ti}$ = 1, sample $w_{ti}$ from the background language model $\xi_1$;
             \item if $s_{ti}$ = 2, sample $w_{ti}$ from the document-specific language model $\xi_{2m}$;
             \item if $s_{ti}$ = 3, sample $w_{ti}$ from the \htopic~$\phi_{z_t}$.
           \end{enumerate}
       \end{enumerate}
   \end{enumerate}
\end{enumerate}
}

\begin{figure*}[t!]
\centering
{
\begin{tabular}{p{16cm}}
[...] in 2006 montreal was named a unesco city of design only one of three cities [...] \\
\emph{[...] in 2006 c7954681 was named a c21786641 c8560 of c2318702 only one of three c8560 [...]} \\ \hline
[...] the biggest sport following in montreal clearly belongs to hockey [...] \\
\emph{[...] the biggest c25778403 following in c7954681 clearly belongs to c10886 [...]} \\
\end{tabular}
}
\vspace{-3mm}
\caption{Text fragments before and after concept identification.}
\label{fig:disambig}
\end{figure*}

\subsection{A 2-level LDA}
\label{sec:lda}
\paragraph{Paragraphs} are modeled as generated from a mixture of three {\it word(-level) topics} (\ltopics) corresponding to each of (1) background, (2) document-specific, (3) ``theme-specific'' language model respectively, exemplified in the left side of Figure \ref{fig:overview}. This modeling is used as a filter -- the (collection-specific) background and the document-specific words are not informative with respect to themes, and are skipped in the next processing step. While only side-effects with respect to the focus of this paper, the background and document-specific language models can be useful: the document-specific word probability distribution can be used to align, cluster or classify documents. We explore this briefly in Section \ref{sec:results}. The \ltopic~mixture follows a Dirichlet distribution with hyperparameter $\gamma$ ($Dirichlet(\gamma)$), while words within a \ltopic~ follow $Dirichlet(\eta)$. 
The topic $s_i$ assigned to word at position $i$ in paragraph $t$ is sampled according to:

\[ p(s_i = l|\s,\wt) \propto g_l(w_{ti}) \mbox{~}\frac{n_{w_{ti}}^l + \eta}{n_*^l + W \eta}\mbox{~}\frac{n_{t-i}^l + \gamma}{n_{t-i}^* + 3\gamma} \] 

\noindent where $W$ is the size of the vocabulary, $n_w^l$ is the number of times word $w$ was assigned to \ltopic~ $l$, $*$ is a wild-card, $n_t^l$ is the number of words in paragraph $t$ assigned to topic $l$, $n_t^*$ is the number of words in paragraph $t$, and $\wt_{-i}$ is the sequence of words $\wt$ minus the word $w_{ti}$ at position $i$. $g_{1..3}$ are normalized coefficients that capture the bias of each word for the three postulated topics, according to their document and collection frequencies\footnote{Background language $g_1(w_{ti}) = \frac{c_{w_{ti}}^P}{c_*^P}$ (generally frequent words); document-specific $g_2(w_{ti}) = \frac{c_{w_{ti}}^d}{c_*^d}$ (common throughout the document, rare outside); topic-carrying  $g_3(w_{ti}) = \frac{c_{w_{ti}}^D}{c_*^D} \times (1 - g_1(w_{ti}))$ (appear throughout the collection, but not frequent within single documents). $P$ is the total number of paragraphs in the set of documents $D$, $d$ is a document (the document being processed), $c_x^y$ is the number of $y$s in which $x$ appears.Because $g_l$ are constant, they will not be affected by marginalizing $\xi$: $p(\wt,\g|\s,\eta) = \int p(\wt,\g|\s,\xi) p(\xi|\eta) d\xi = \int p(\g|\s) p(\wt|\s,\xi) p(\xi|\eta) d\xi = \prod_{i} g_{s_i}(w_{ti}) \int p(\wt|\s,\xi) p(\xi|\eta) d\xi$. These factors are normalized such that for each word they sum to 1.}.

\paragraph{Documents} are generated from a mixture of $K$ {\it themes} (\htopics) following $Dirichlet(\lambda)$. The \htopic~of a paragraph is determined based on the subset of its {\it theme-specific words}, as determined by the previous step. The words in a topic follow $Dirichlet(\beta)$. The right side of Figure \ref{fig:overview} shows an example of the \htopics~within a document. \footnote{ The \htopics~induced by the model are nameless, we name them in the figure to illustrate more clearly the point.} Contiguous paragraphs that express the same \htopic~form a segment, and through this can be aligned across documents and languages. Paragraph \htopics~are sampled as:

{\[p(z_t=j|\z,\wt,\lambda,\beta) \propto \frac{n_{w_{ti}}^j + \beta}{n_*^j + W \beta} \frac{n_{t-i}^j + \lambda}{n_{t-i}^* + K\lambda}  \]}






\subsection{Modeling the sequence of topics/themes}
\label{sec:sequence}
The sequence of themes within a document is modeled through an HMM. The transition probabilities are modeled by a Dirichlet distribution \cite{beal02}: the transition probabilities from a state $z_t = j$ at time $t$ can be interpreted as mixing proportions for the state at time $t+1$: $\pi_j = \{\pi_{j1}, ..., \pi_{jK}\}$. The {\bf persistence of states} is encouraged by increasing the probability of self-transitions using a ``sticky'' parameter $\kappa$ \cite{beal02,fox10}: $\pi_i | \alpha,\kappa \sim Dirichlet(\alpha+\kappa\delta_i) $, where $\kappa>0$ is added to the $i^{th}$ component of the parameter vector $\alpha$\footnote{\newcite{fox10} model the transitions through a Dirichlet process to allow the model to infer the number of states.}. \htopics~ are sampled as:

\begin{tabular}{lll}
\multicolumn{3}{l}{$ p(z_t = j | z_{-t},\alpha,\kappa,\lambda) \propto \qquad$} \\ [3mm]
 ~~~ & $ \frac{n_{-t,j}^{(d)} + \lambda}{\sum_l n_{-t,l}^{(d)} + K\lambda}$ & \parbox[c]{4cm}{\small topic mixing \newline proportion} \\ [5mm]
 ~~~ & $ \frac{n_{z_{t-1}j}^{-t} + \alpha_j + \kappa*\delta(z_{t-1},j)}{n_{z_{t-1}*}^{-t} + \sum_x\alpha_x + \kappa}$  & \parbox[c]{7cm}{\small transition from \newline previous paragraph}\\ [5mm]
 ~~~ & $\frac{n_{jz_{t+1}}^{-t} + \alpha_j + \kappa*\delta(j,z_{t+1})}{n_{j*}^{-t} + \sum_x\alpha_x + \kappa} $ & \parbox[c]{7cm}{\small transition to next \newline paragraph}\\ [3mm]
\end{tabular}


\vspace{0.2cm}
\noindent where {$ \delta(i,j) = \left\{ \begin{array}{rl} 1 & i == j \\ 0 & i \ne j  \end{array}  \right. $} \begin{small}(Kronecker's delta)\end{small}.

\vspace{0.5cm}
After sampling the \htopic~ for paragraph $t$, the \htopics~of the topic-carrying words in $\wt$ (assigned \ltopic~3) are reassigned to the paragraph's \htopic, reflecting the assumption that there is one \htopic~per paragraph.

\subsection{Language-independent concepts}\label{sec:disambig}
\label{sec:concepts}
To bridge languages, and address (at least partly) the issues of multi-word expressions, synonymy, polysemy, we introduce concept annotations, exemplified in Figure \ref{fig:disambig}. To identify concepts:

\begin{itemize}
 \item locate possible concept lexicalizations extracted from Wikipedia in texts;
 \item for each lexicalization, find all candidate concepts it could refer to;
 \item disambiguate among possible candidate concepts.
\end{itemize}



For disambiguation, the text is represented as a complete $n$-partite graph $G = (V_1, .. V_n,E)$. Each partition $V_i$ corresponds to a (possibly multi-word) term $t_i$ in the text, and contains as vertices all concepts $c_{i_j}$ that may be expressed by $t_i$, found in the first step. Each vertex from a partition is connected to all vertices from the other partitions (making the n-partite graph complete) through edges $e_{v_{i},v_{j}} \in E$ weighted by $w_{v_{i},v_{j}}$. The weights are learned from Wikipedia's link structure, using as features several measures: shared outgoing links between the articles corresponding to the two concepts, shared categories and their specificity; preference of each concept for the other concept's expression through the anchor in the current text.

\begin{figure}[h]
  \centering
  \includegraphics[scale=0.15]{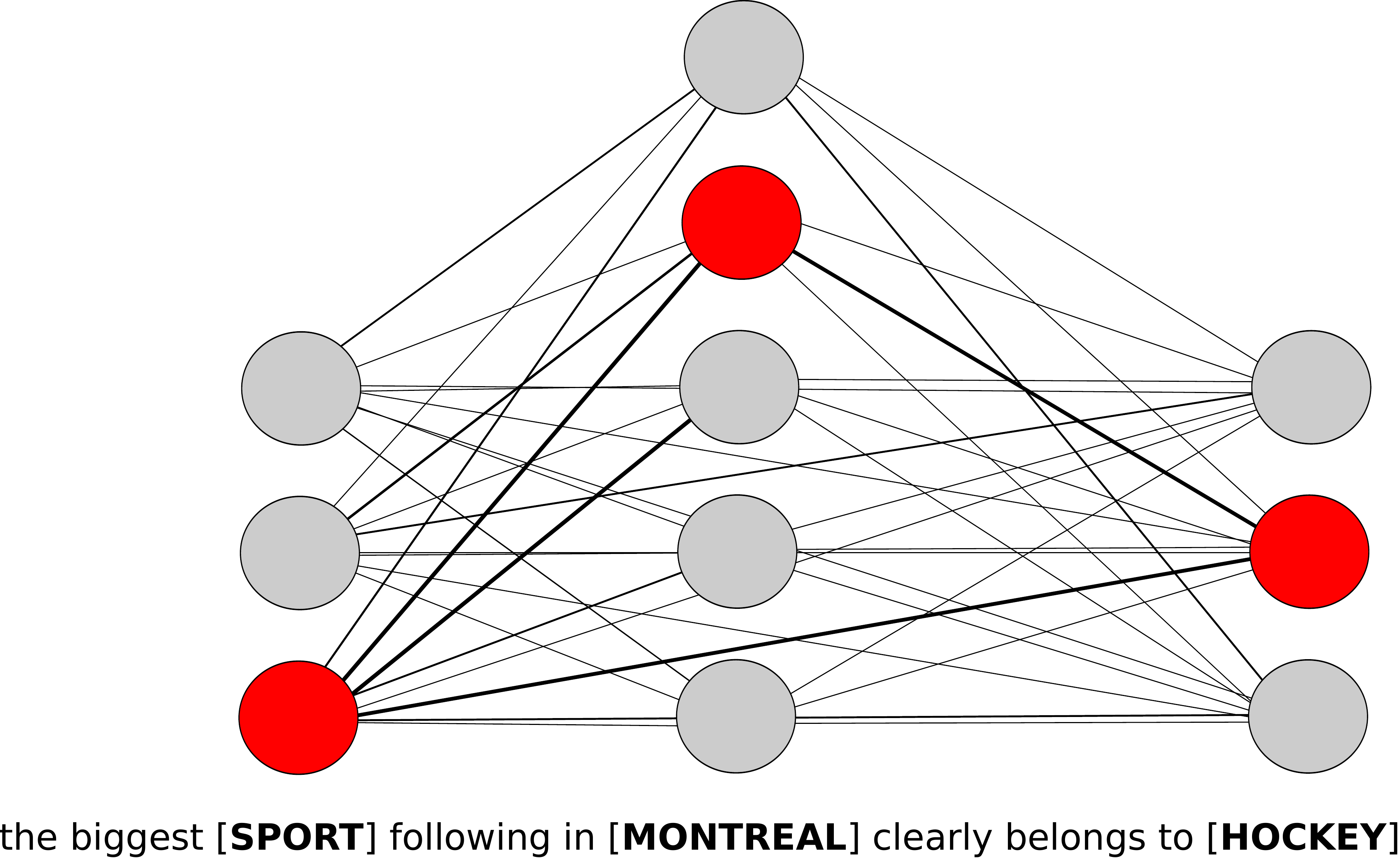}
  \caption{Weighed $n$-partite graph. Edge weights are represented by thickness.}
  \label{fig:graph}
\end{figure}

The disambiguated concepts are the nodes in the maximum edge-weighted clique in this graph, and their expression in the text is replaced with the corresponding unique (and language independent) concept ID, resulting in the concept-annotated texts illustrated in Figure \ref{fig:disambig} \cite{fahrni11}. This concept identification method scored high on both ACE 2005 (72.7\% F-score) and in the Entity Linking TAC 2011 task.

The texts with concept identifiers are processed with the previously described topic model. To take advantage of the relations between concepts, we introduce a factor $s_{wkn}$, to give a boost to topics that are favoured by the concepts $w_r \in \mathcal{R}_{w_{ti}}$ directly connected to $w_{ti}$ in the concept network (for $w_{ti}$ that represent a concept):

{\[ s_{wkn}(w_{ti}|z_i) = \frac{1}{| \mathcal{R}_{w_{ti}}|}\sum_{w_r \in \mathcal{R}_{w_{ti}}} \frac{n_{w_r}^{z_i}}{n_{w_r}^{*}} \]}

\noindent where 
\[\mathcal{R}_{w_{ti}} = \{ w_r | (w_r,w_{ti}) \mbox{ is an edge in the network of concepts}\}\]

\noindent $n_{w_r}^{z_i}$ is the number of times concept $w_r$ was assigned \htopic~$z_i$ (excluding the current occurrence), and $n_{w_r}^{*}$ is the number of times concept $w_r$ was assigned any \htopic.

\subsection{Topic assignment}

To make the final assignments of \htopics~to paragraphs in the test data, we use the word probabilities under both \ltopics~and \htopics, and the transition probability distributions for \htopics~induced as a result of the iterative sampling process in the HMM framework. States correspond to \htopic~assignments to paragraphs. The final assignments of \htopics~to the current document is determined through a Viterbi algorithm on the HMM with the parameters described above.




\section{Experiments}\label{sec:experiments}

\paragraph{Data} 

The data consists of a collection of articles about cities in English and French presented in \cite{chen09}.\footnote{\url{http://groups.csail.mit.edu/rbg/code/mallows}.} Table \ref{tab:data} presents the data statistics. These documents have structure -- sections, such as History, Culture, Transportation -- which is what we try to recreate through this process\footnote{\newcite{yahyaei11}'s data did not fit this structure, and we could not use it to test this method.}. The English data was manually annotated with a ``clean'' set of headings, to allow mapping of sections that have the same topic but slightly different headings (e.g. Culture and arts/Culture). This manual annotation process revealed 18 topics that appear in more than one document. To evaluate the cross-lingual topic alignment, the French paragraph labels were (manually) translated to English, if a corresponding label existed on the English side -- e.g. {\it histoire} was translated to {\it history}; {\it voir aussi} was translated as {\it further reading}, its English side counterpart (instead of the more literal {\it see also}). Labels that had no correspondent in English were not translated.

\begin{table}[h]
\centering
\begin{tabular}{l|l|l|l|l}
Corpus & Language & Docs & Pars & Vocab. \\ \hline
CitiesEn & English & 100 & 6670 & 42,603 \\ \hline
CitiesFr & French & 100 & 4074 & 31,487 \\
\end{tabular}
\vspace{-3mm}
\caption{Data statistics}
\label{tab:data}
\end{table}

\vspace{-3mm}

\paragraph{Modeling parameters}

The parameters of the model reflect the differences between the two languages data (e.g. English has 1.5 the number of paragraphs that French does, and also a larger vocabulary): 
\begin{itemize}
 \item for word-topic distributions: values $<1$ to bias towards distributions that favour high probabilities for a small set of words for both \ltopics~ and \htopics: $\beta = \eta = \frac{W}{100000}$ ($W$ is the size of the vocabulary in English and French respectively); 
 \item for topic mixtures: values $>1$ to produce balanced topic mixtures within paragraphs (for \ltopics) $\gamma = \frac{W}{|P|}$ ($|P|$ is the number of paragraphs), and within documents $\lambda = 50/K$ (for both languages, following \cite{griffiths04}); 
 \item for state transitions: a high $\kappa = 1000$ to encourage state persistence; and low $\alpha = 0.01$ to bias the state transition model to prefer a small number of possible succeeding topics. 
\end{itemize}

\subsection{Monolingual alignment}

The performance of the model is evaluated based on the assigned \htopics~on the task of cross-document alignment \cite{chen09}. Reference topic assignments are the section labels chosen by their authors. Following \cite{chen09} we compute:

\[ Rec = \frac{\sum_{h\in H}max_{k \in K}(overlap(h,k))}{P} \]

\[ Prec = \frac{\sum_{k \in K}max_{h \in H}(overlap(h,k))}{P} \]

\noindent where $P$ is the number of paragraphs, $H$ is the set of section headings, $K$ is the set of automatically assigned \htopics (themes), and $overlap(h,k)$ is the number of paragraphs with section heading $h$ and automatically assigned topic $k$.

\subsubsection{Settings}

\paragraph{Baselines} Gruber et al.'s \shortcite{gruber07} {\it HTMM} models a document as a sequence of sentences. All words in a sentence have the same topic, and a binomial transition parameter determines whether the next sentence has the same topic as the current one. \newcite{chen09} propose {\it GMM}, a global model for documents in a collection. They assume that the documents in the collection share the same topics, which are similarly distributed within the documents. This is captured by modeling the mixture of topics as a distribution over permutations of a topic ordering.

\paragraph{Our variations} {\it 2LDA} is the two level LDA processing as described in Section \ref{sec:lda}, on the texts with concept annotations. {\it 2LDA\_HMM} adds the HMM to 2LDA to model the transition between paragraphs (Section \ref{sec:sequence}). {\it 2LDA\_c} adds the score based on concept relations to 2LDA. {\it 2LDA\_c\_HMM} adds the score based on concept relations to 2LDA\_HMM. 

{\it 2LDA\_HMM no concepts} is the best performing configuration without concept annotations.


\subsubsection{Results}\label{sec:results}

\paragraph{Language models}

The first result consists of the language models induced by the system. Figure~\ref{fig:models} shows examples of the general (background), document-specific (for the English and French Wikipedia articles about the city of Montreal) and two \htopic~language models. The top section of the figure shows the background and document-specific language models for the English and French data (shown side by side because the documents refer to the same cities). The background language model has captured what is usually included in a list of stopwords. This is an advantage, as stopwords may be collection-specific. The document-specific language model include words expected to be associated with the respective article titles. Sample \htopic~language models are shown separately for each language. 

\begin{figure}[t]
\centering
{\footnotesize
\renewcommand{\arraystretch}{0.7}
\begin{minipage}[t]{16cm}
\parbox[t]{3cm}{
\begin{tabular}{lll}
\multicolumn{3}{l}{{\bf Background}} \\ \hline 
en & fr & combined \\ \hline
the & de & the \\
of & la & of \\
and & le & and \\
in & et & in \\
to & les & de \\
a & des & a \\ \\
\end{tabular}
} 
~\\
\parbox[t]{8cm}{
\begin{tabular}{llp{2cm}l}
\multicolumn{4}{l}{{\bf Document-specific: Montreal}} \\ \hline
en (\#100) & fr (\#12) & en (concepts) & fr (concepts) \\ \hline
montreal & motreal & {\it Montreal} & {\it Montreal}\\
canada & quebec & {\it Canada} & {\it Quebec} \\
montreals & canada & {\it Melbourne} & {\it Canada}\\
de & l\^{i}le & {\it Of Montreal} & l'ile \\
french & fran\c{c}ais & {\it North America} & fran\c{c}ais \\
montr\`{e}al & fran\c{c}aise & {\it Saint Lawrence River} & {\it language} \\ \\
\end{tabular}
}
~\\ 
\parbox[t]{8cm}{
\begin{tabular}{lll}
\multicolumn{3}{l}{{\bf HL-topic \#1}} \\ \hline
en & fr & multilingual \\ \hline
cricket & football & {\it team} \\
basketball & tennis & {\it stadium} \\
baseball & club & {\it sport}\\
usa & clubs & {\it website} \\
tennis & coupe & {\it Olympic games}\\
hockey & stade & {\it photography} \\ \\
\end{tabular}
}
~\\
\parbox[t]{8cm}{
\begin{tabular}{lll}
\multicolumn{3}{l}{{\bf HL-topic \#2}} \\ \hline
en & fr & multilingual \\ \hline
terminal & trains & {\it bus} \\
airlines & a\'{e}roports & {\it passenger}\\
airports & passagers & {\it train station}\\
commuter & stations & {\it rapid transit} \\
terminals & gare & {\it video}\\
cargo & route & {\it train}\\
\end{tabular}
}
\end{minipage}
}
\caption{Ordered sample of words from the background, document-specific and \htopic~language models. The terms in italics are concepts, and we provide their English names for legibility.}
\label{fig:models}
\end{figure}

Terms in the text are replaced with language independent concept IDs. We combine the English and French data, and apply the topic model to this data. Samples of the resulting \htopics~ are included in Figure \ref{fig:models}  -- columns labeled ``multilingual''. The terms in {\it italics} represent concepts -- for readability replaced with the name of the corresponding article in the English Wikipedia. We have also performed experiments with building separate topics for English and French, while sharing only the language-independent concepts (i.e. after building the topics for English, we use the English topic-annotated data as observed data, and adjust the prior computed from this data while iterating over the French data). The results were very similar, and for reasons of space are not included.

The background and document-specific language models are ``side-effects'' of the process of cross-document and cross-language alignment. However, they can be useful. The document-specific topics could be used for (cross-language) document clustering. We took the top 20 words and concepts ranked based on tf-idf, as well as the top 20 terms in the document-specific topics for each of the 200 documents (100 for each language), and used the cosine similarity to align the documents in the two languages (i.e., to pair up the documents in English and French that were about the same city). The baseline -- using tf-idf scored words -- is already high (94\%), due to shared city and country names. Both topic probabilities and tf-idf scored concepts perform perfectly (100\%). For the task and data described in this paper, document level alignment performed using document-specific topics can be compounded with the theme-based paragraph alignment to produce paragraph alignment between pairs of documents. We plan to study the usefulness of the document-specific topics in a more difficult environment.

\paragraph{Alignment}

\begin{figure*}[t!]
\centering
\begin{minipage}[t]{16cm}
\resizebox{7.5cm}{!}{\includegraphics{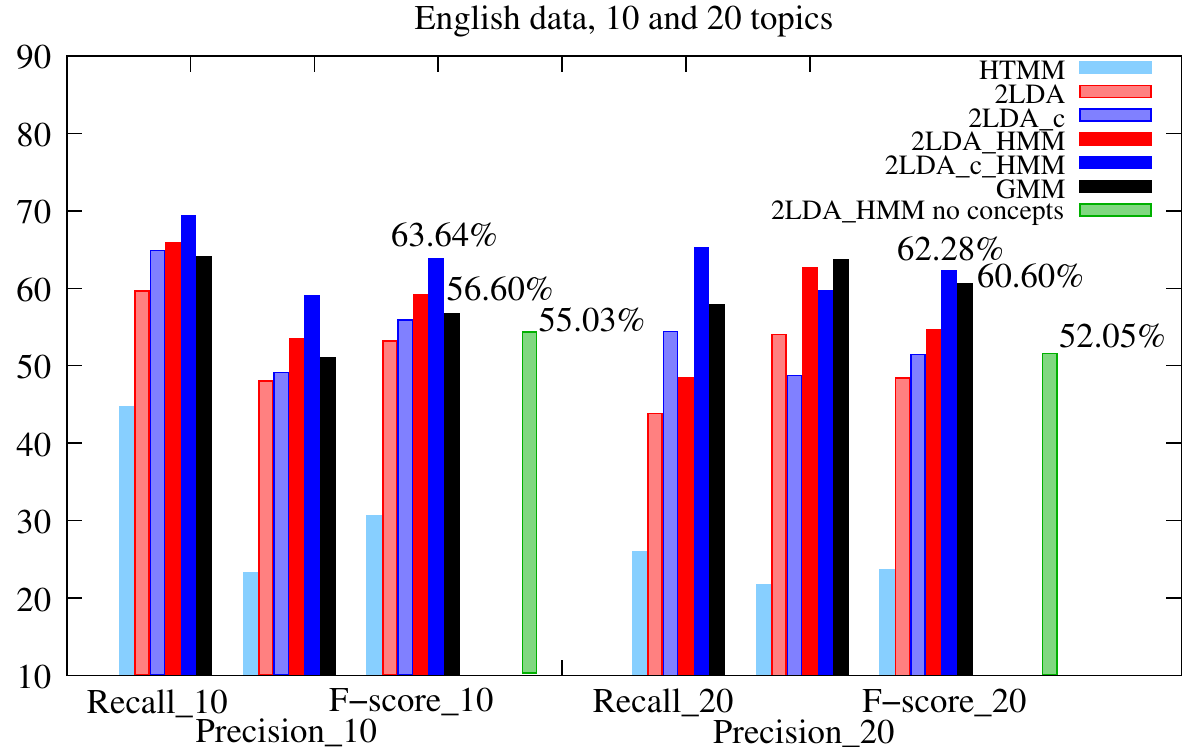}}
\hfill
\resizebox{7.5cm}{!}{\includegraphics{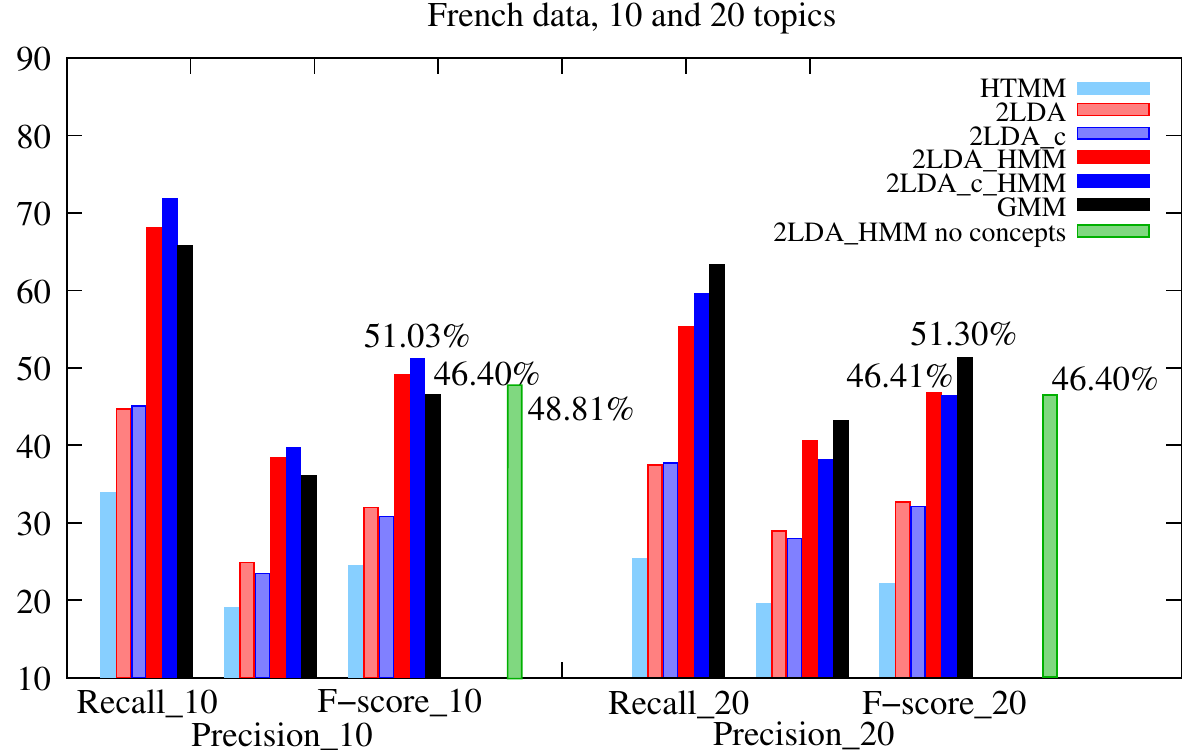}} 
\end{minipage}
\vspace{-5mm}
\caption{Alignment evaluation in terms of Precision, Recall and F-score.}
\label{fig:results}
\end{figure*}

Figure \ref{fig:results} presents the results in terms of precision, recall and F-score for the settings describe above to reveal the contribution of the HMM and the concept annotations: The results on the English data, for both 10 and 20 \htopics, show a nice progression of the results as we add more information in the model. Adding HMM leads to better results, as it allows transition probabilities -- and through this the context -- to influence a \htopic~assignment for each paragraph in the document. Adding concepts and their relations in the model leads to increase in recall, which is again expected since this will lead to more links between terms in different parts of the document. The results are better than previous work, with the exception of the situation when we use 20 \htopics~on the French data. 

\subsection{Cross-language alignment} 
Cross-language alignment is performed as for monolingual alignment, but we compare topic assignments across languages. 

\subsubsection{Settings}
\paragraph{Baselines} {\it Baseline~1} is based on similarity between paragraphs computed using translation probabilities from a translation table. {\it Baseline~2} computes paragraph alignment using a concept-based representation of each paragraph with concepts weighted according to their tf-idf score and a cosine similarity metric. Both baselines are computationally intensive because of all the pairwise comparisons between paragraphs.

\paragraph{Our approach} {\it 2LDA\_c\_HMM} We experimented with concatenating the corpora in the two languages and inducing topics from the union, or alternatively using the distributions induced on one as priors when processing the other. The results are not significantly different.

\subsubsection{Results}
 Cross-lingual precision, recall and F-score computed over the entire (bilingual) collection are presented in Table \ref{tab:xalign}, for 10 and 20 topics.

\begin{table}[h]
\centering
\begin{tabular}{@{}l|lll@{}}
Method & P & R & F\\ \hline
Baseline 1: transl. table & 79.06 & 30.07 & 43.56 \\
Baseline 2: concepts & 81.15 & 29.09 & 47.10 \\ \hline
2LDA\_c\_HMM, 10 topics & 47.14 & 67.90 & 55.65 \\ 
2LDA\_c\_HMM, 20 topics & 46.85 & 63 & 53.74 \\ 
\end{tabular}
\caption{Cross-language alignment results.}
\label{tab:xalign}
\end{table}

The high precision results for the two baselines are due in large part to singleton clusters, many of which are in fact correct. Building only singleton clusters leads to an F-score of 35.13.


\section{Conclusions}
We have explored knowledge-enhanced -- through concept annotations -- topic models. The annotations not only help build cross-lingual topic models, but also deal with multi-word terms, polysemy and synonymy. The links between concepts are also beneficial by influencing the topic distribution and sequencing probabilities. The results are good despite the fact that we rely on an automatically built resource, and a concept identification step -- itself an open research problem. We plan to investigate a joint approach to topic modeling and concept identification.

We have augmented the traditional HMM model with language models within a paragraph, and a Dirichlet distribution at the level of state (high-level topic) transitions. Modeling a text on two levels gives a layered view, which can be used to structure the commonly used bag-of-word representation of texts.
Separating words that contribute to the topic segmentation from those that pertain to a background language or document-specific model helps the system focus on the most relevant terms for segmentation. This could become an advantage when aligning corpora in different languages -- one need only establish parallelism between topic-specific terms.

\bibliographystyle{acl}
\bibliography{lit,extras} 

\end{document}